\documentclass{article}
\pdfoutput=1

\usepackage[preprint, nonatbib]{Styles/neurips_2024}

\usepackage[utf8]{inputenc} % allow utf-8 input
\usepackage[T1]{fontenc}    % use 8-bit T1 fonts
\usepackage{hyperref}       % hyperlinks
\usepackage{url}            % simple URL typesetting
\usepackage{booktabs}       % professional-quality tables
\usepackage{amsfonts}       % blackboard math symbols
\usepackage{nicefrac}       % compact symbols for 1/2, etc.
\usepackage{microtype}      % microtypography
\usepackage{xcolor}         % colors
\usepackage{graphicx}       % for including images
\usepackage{amsmath} 
\usepackage{mathrsfs}
\usepackage{multirow} 
\usepackage{authblk}

\definecolor{darkgreen}{rgb}{0.439, 0.678, 0.278}
\definecolor{grayishblue}{rgb}{0, 0.439, 0.753}

\title{Searching Priors Makes Text-to-Video Synthesis Better}

\author[1]{\textbf{Haoran Cheng}}
\author[2]{\textbf{Liang Peng}}
\author[1]{\textbf{Linxuan Xia}}
\author[3]{\textbf{Yuepeng Hu}}
\author[1]{\\\textbf{Hengjia Li}}
\author[4]{\textbf{Qinglin Lu}}
\author[1]{\textbf{Xiaofei He}}
\author[5]{\textbf{Boxi Wu}}
\affil[ ]{
    $^1$ State Key Lab of CAD\&CG, Zhejiang University
    $^2$FABU Inc. 
}
\affil[ ]{
    $^3$Ningbo Port
    $^4$Tencent Data Platform 
    $^5$College of Software, Zhejiang University 
}
\affil[ ]{\texttt{\{haorancheng, pengliang, xialinxuan, wuboxi\}@zju.edu.cn, huyp@nbport.com.cn}}
\affil[ ]{\texttt{lihengjia98@gmail.com, qinglinlu@tencent.com, xiaofeihe@cad.zju.edu.cn}}
\affil[ ]{}
\affil[ ]{
    \url{https://hrcheng98.github.io/Search_T2V/}
}

\begin{document}

\maketitle

\begin{figure}[h]
  \centering
  \includegraphics[width=1\textwidth]{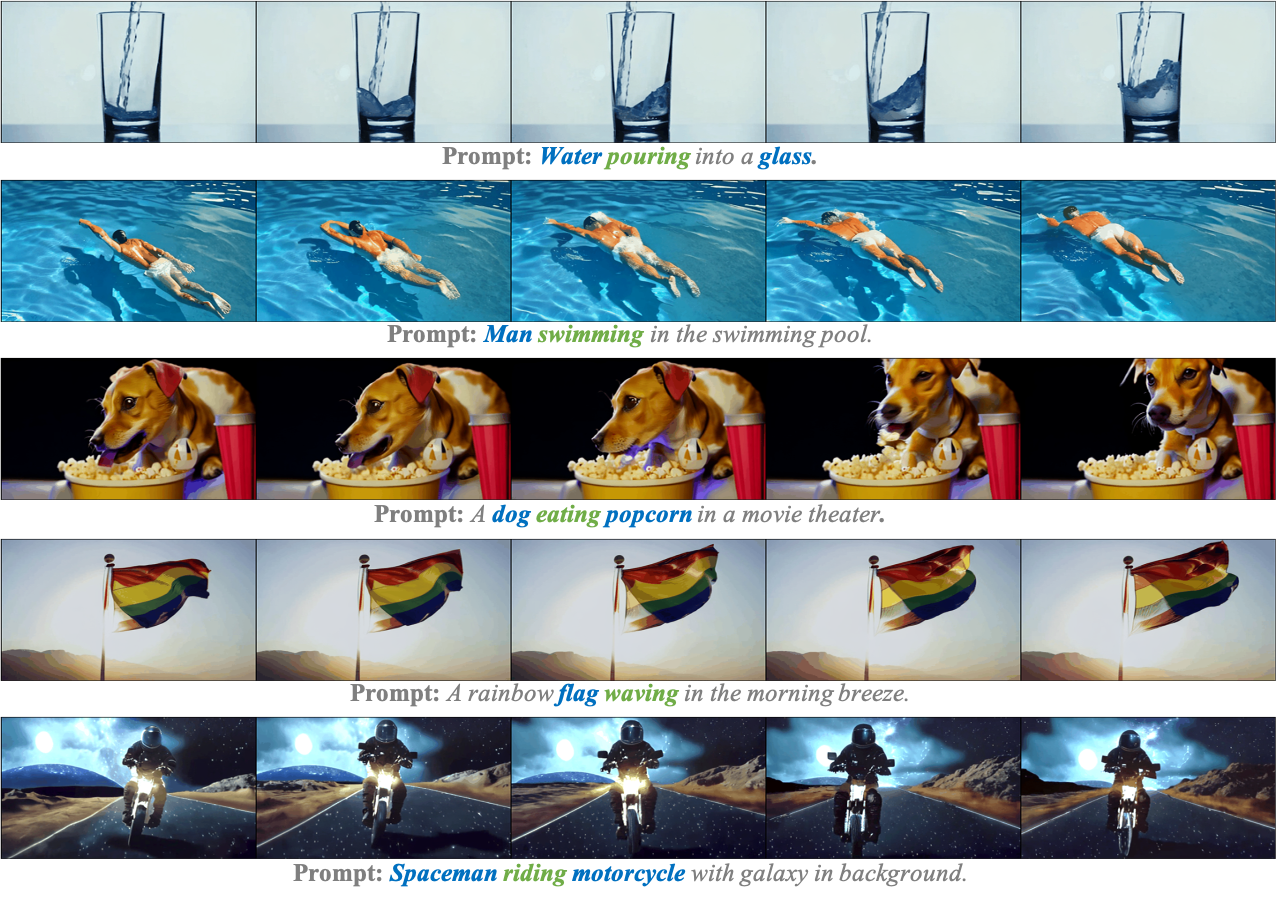} % Adjust the width as needed
  \caption{Samples produced by the proposed text-to-video method. 
  The highlighted words of prompts are core semantic units extracted by the proposed method, 
  guiding the search for prior knowledge during video generation. 
  \textcolor{darkgreen}{\textbf{\textit{Green}}} highlights indicate the motion semantics of the motion,
  while \textcolor{grayishblue}{\textbf{\textit{blue}}} ones indicate the objects directly related to the motion.
  Please refer to \href{https://hrcheng98.github.io/Search_T2V/}{\textbf{\textcolor{blue}{\textit{Website}}}} for the best view.}
  \label{fig:sample}
\end{figure}

\begin{abstract}
    Significant advancements in video diffusion models have brought substantial progress to the field of text-to-video (T2V) synthesis.
    However, existing T2V synthesis model struggle to accurately generate complex motion dynamics, 
    leading to a reduction in video realism. 
    One possible solution is to collect massive data and train the model on it, 
    but this would be extremely expensive.
    To alleviate this problem, in this paper, we reformulate the typical T2V generation process as a search-based generation pipeline. 
    % Instead of collecting and training on massive data, 
    Instead of scaling up the model training, we employ existing videos as the motion prior database.   
    Specifically, we divide T2V generation process into two steps: 
    (i) For a given prompt input, we search existing text-video datasets to find videos with text labels that closely match the prompt motions. 
    We propose a tailored search algorithm that emphasizes object motion features. 
    (ii) Retrieved videos are processed and distilled into motion priors to fine-tune a pre-trained base T2V model, followed by generating desired videos using input prompt. 
    By utilizing the priors gleaned from the searched videos, we enhance the realism of the generated videos' motion. 
    % Moreover, we propose a search algorithm tailored for video generation that emphasizes object motion features. 
    All operations can be finished on a single NVIDIA RTX 4090 GPU.
    We validate our method against state-of-the-art T2V models across diverse prompt inputs.
    % We validate our method on standard video generation benchmarks and human evaluations. 
    The code will be public.
\end{abstract}

\section{Introduction}

    % Video synthesis is a challenging and long-studied task, 
    % which combines the challenging components of language modeling and image generation.
    % A large of number
    In recent years, diffusion models~\cite{diffusionbeatgan, ddpm, ldm} have demonstrated impressive capabilities to create high-quality visual content, capturing the interest and enthusiasm of both the general public and the academic community. 
    Recently, diffusion models have brought forward milestone improvement for video synthesis~\cite{vdm, imagenvideo, makeavideo, tuneavideo}. 
    % Given the text prompts, users 
    Most text-to-video (T2V) models usually leverage successful pre-trained text-to-image (T2I)~\cite{ldm, ramesh2022hierarchical, saharia2022photorealistic, gu2022vector} models by extending the architecture from 2D to 3D and training them on public video datasets.
    
    However, the current video synthesis models often generate motion dynamics that are inconsistent with the real world. 
    We identify the issues as follows: 
    (1) Unnatural object movements, where objects may move at unusual speeds or in unrealistic ways; 
    (2) Inconsistent object spatial relationships, where object positions may change illogically; 
    (3) Improbable interactions, including unlikely object interactions or collisions that rarely happen in reality. 
    These discrepancies reduce the realism of generated content and highlight the need for improvements in motion fidelity. 
    % Increasing training dataset size and model volume may potentially alleviate these issues, but given the diversity and complexity of real-world motion dynamics, this approach would be extremely expensive.

    One potential method to alleviate these challenges could be increasing the training dataset size and scaling up the model capacity.
    % Increasing the size of the training dataset and the model capacity could potentially mitigate these challenges.
    However, given the vast diversity and complexity inherent in real-world motion dynamics, this approach would be prohibitively expensive. 
    The complexity of accurately capturing and modeling real-world movements necessitates extensive model training on large-scale video datasets.
    This makes it an impractical solution for many research and industry contexts, particularly in scenarios where budgets and resources are constrained.
    
    To generate high-quality videos with realistic motions in low cost, we introduce an intuitive approach. 
    % To generate high-quality videos with realistic motions, we introduce a cost-effective and intuitive approach. 
    The internet today is replete with a wealth of real video resources that encompass diverse visual motion information, offering an abundance of reference sources. 
    This inspires us to transform the conventional large-scale training and generating approach into a combination of searching and generating, to eliminate the high costs associated with large-scale video model training. 
    Namely, we reformulate the typical text-to-video pipeline as a search-based method. 
    Instead of training the model on vast amounts of video data to learn realistic motions, we extract motion semantics from the input prompt and search existing real videos for obtaining similar motion characteristics. 
    These videos serve as motion references and are integrated into the base T2V model to enhance the realism of the synthesized videos. 
    Specifically, we first search existing text-video datasets~\cite{wang2019vatex, webvid, hdvg} to identify text labels that closely align with the input prompt, obtaining corresponding real videos. 
    We then distill the dynamic information~\cite{vmc} from these real videos and embed it as motion priors into the generator~\cite{show1} for video synthesis through video motion customization~\cite{zhao2023motiondirector,customizeavideo, vmc, materzynska2023customizingmotion, zhang2023motioncrafter}.
    
    Moreover, we propose a video retrieval module tailored to search object motion characteristics from video datasets. 
    The proposed module analyzes the input prompt 
    % to extract motion features (e.g., verbs) and object attributes (e.g., nouns), 
    to extract motion (e.g., verbs) and object (e.g., nouns) semantic features.
    % focusing on matching \textcolor{red}{dataset} containing motion features from the prompts. 
    % The algorithm emphasizes the extraction of object motion characteristics within the video editing model.
    We match the extracted prompt semantics with the semantics of the text in the dataset, selecting the text-video pair whose motion information is the most similar to the original input, which could provide the most related priors for subsequent fine-tuning.
    
    In summary, our contributions are listed as follows:
    \begin{itemize}
        \item We formulate a new text-to-video generation pipeline, 
         which performs a search-based generation process by obtaining valuable motion information under low costs.
        \item We introduce a video retrieval module focusing on searching object motion characteristics from existing video datasets. 
        It aims to provide closely aligned videos with the input prompt.
        % \item We evaluate our model on standard video generation benchmarks and obtain significant improvements.
        \item We validate our method against state-of-the-art T2V models across diverse prompt inputs.
    \end{itemize}

\section{Related Works}

% \subsection{Video Diffusion Model}
% \subsection{Video Synthesis}
\subsection{Text-to-Video Synthesis}
    % The text-to-video synthesis, a challenging and long-studied task, involves producing videos that align with the language modeling.
    The text-to-video synthesis, a challenging task, involves producing videos that align with the input prompt.
    Early works use a wide range of generative models,
    including generative adversarial network (GAN) ~\cite{gan, gan_vondrick2016generating, gan_saito2017temporal, gan_tulyakov2018mocogan, gan_tian2021good},
    autoregressive models~\cite{auto_srivastava2015unsupervised, yan2021videogpt} and implicit neural representations~\cite{stylegan, implicit_tian2021good}.
    Recently, the advancement of diffusion models~\cite{diffusionbeatgan, ddpm, ldm} in the text-to-image field brings great improvements to text-to-video tasks.
    A popular fashion for text-to-video is to inflate a pre-trained T2I model by inserting spatial-temporal 3D convolutions and then conducting training on video data.
    Several works~\cite{vdm,videoldm,makeavideo,imagenvideo} have elucidated the effectiveness of this approach.
    Recent studies~\cite{svd, videocrafter2} demonstrate a data-centric perspective technique to enhance the performance of text-to-video models.
    However, most methods mentioned above require expensive computing costs for training and still exhibit poor performance when generating complex dynamics in the real world. 

\subsection{Video Editing}
    
    % \paragraph{Video-to-video Editing.} 
    Video editing aims to create videos that follow the provided editing instructions (e.g., text and images) while preserving the original characteristics of the source video.
    Some methods take a training-free approach~\cite{meng2021sdedit, bartal2024lumiere} to address the challenge of lacking annotated target video data, achieving promising results. 
    % The lack of annotated target video data for training video editing models is a common challenge, and SDEdit 
    % ~\cite{sdedit} proposed a natural solution to adopt a training-free approach to solve this challenge.
    % Recent studies[Lumiere, SORA] follow this technique. 
    Another popular method involves injecting information about the input or generated video from keyframes through cross-attention~\cite{zeroshot_offtheshelf, renderavideo, tokenflow, t2vzero, ceylan2023pix2video}.
    Some methods~\cite{structure_content, liang2023flowvid} denoise original videos by extracting essential features, like depth maps or optical flow which should be retained in the edited video. 
    Additionally, some works~\cite{tuneavideo, peng2024smoothvideo, neuralvideofieldsediting} invert the input video using the input caption to generate new videos by using inverted noise and a caption describing the target video.
    
\subsection{Video Motion Customization}
    Video customization task aims to customize the video synthesis model to adapt the original output to a personalized concept, by adjusting the pre-trained weights. 
    Early works~\cite{animageworthoneword, dreambooth, multiconcept} introduce this task into text-to-image models to perform appearance customization.
    Building on these insights, recent research works~\cite{zhao2023motiondirector, vmc, customizeavideo, customizingmotion, dreamvideo} focus on customizing the motion based on reference videos.
    % Customizing Motion learns a new motion with a special token from a set of few exemplar videos which indicate the desired target motion. 
    MotionDirector~\cite{zhao2023motiondirector} applied a dual-path architecture and an appearance-debiased temporal training objective to customize motion.
    VMC~\cite{vmc} leverages the residual vectors between consecutive frames as motion vectors to guide the motion distillation objective.
    DreamVideo~\cite{dreamvideo} specially designs adapters over the pre-trained temporal attentions conditioned on one frame to decompose pure motion from its appearance.
    Customize-A-Video~\cite{customizeavideo} filters spatial information more relevant to appearance and employs a temporal LoRA~\cite{hu2021lora} to learn the motion from reference videos.
    Different from video editing, the video motion customization task focuses on learning the motion information in the source video, where the structure and appearance can be changed.

\section{Preliminaries}

\subsection{Denoising Diffusion Probabilistic Models (DDPM)}
DDPM \cite{ddpm} learns the data distribution $q(x_0)$ through forward and reverse Markov processes. Given a variance schedule ${\beta_{t}}{t=1}^T$, noises are added to $x_0$ in the forward process, where the transition is:
\begin{equation}
q(x_t|x{t-1}) = \mathcal{N}(x_t; \sqrt{1-\beta_t} x_{t-1}, \beta_t \mathbb{I}).
\end{equation}

Using Bayes' rule, $q(x_t|x_0)$ is $\mathcal{N}(x_t; \sqrt{\Bar{\alpha}t} x_0, (1 - \Bar{\alpha}t) \mathbb{I})$, with $\Bar{\alpha}t = \prod{s=1}^t \alpha_s$ and $\alpha_t = 1 - \beta_t$. In the reverse process, starting from $p(x_T) = \mathcal{N}(0, \mathbb{I})$, the sampling is done via:
\begin{equation}
p\theta(x{t-1}|x_t) = \mathcal{N}(x_{t-1}; \mu_\theta(x_t, t), \Sigma_\theta(x_t, t)).
\end{equation}

The parameters $\theta$ are trained by maximizing the variational lower bound on the KL divergence, with an objective simplified to:
\begin{equation}
\mathbb{E}_{x, \epsilon \sim \mathcal{N}(0, 1), t} \left[ | \epsilon - \epsilon\theta(x_t, t) |^2_2 \right].
\end{equation}

\subsection{Video Diffusion Model}
Video diffusion models train a 3D U-Net to denoise from a randomly sampled sequence of Gaussian noises to generate videos, guided by text prompts. 
The 3D U-net basically consists of down-sample, middle, and up-sample blocks. Each block has several convolution layers, spatial transformers, and temporal transformers.
%as shown in Fig. 3. 
The 3D U-Net $\epsilon_\theta$ and a text encoder $\tau_\theta$ are jointly optimized by the noise-prediction loss, as detailed in \cite{diffusionbeatgan}:
\begin{equation}
\mathcal{L} = \mathbb{E}_{z_0, y, \epsilon \sim \mathcal{N}(0, I), t \sim \mathcal{U}(0, T)} \left[ \left\| \epsilon - \epsilon_\theta (z_t, t, \tau_\theta(y)) \right\|_2^2 \right],
\end{equation}
where $z_0$ is the latent code of the training videos, $y$ is the text prompt, $\epsilon$ is the Gaussian noise added to the latent code, and $t$ is the time step. 
The noised latent code $z_t$ is determined as follows:
\begin{equation}
z_t = \sqrt{\bar{\alpha}_t} z_0 + \sqrt{1 - \bar{\alpha}_t} \epsilon, \quad \bar{\alpha}_t = \prod_{i=1}^{t} \alpha_t,
\end{equation}
where $\alpha_t$ is a hyper-parameter controlling the noise strength.

%%%%%%%%%%%%%%%%%%%%%%%%%%%%%%%%%%%%%%%%%

\section{Methodology}
\begin{figure}[h]
  \centering
  \includegraphics[width=1\textwidth]{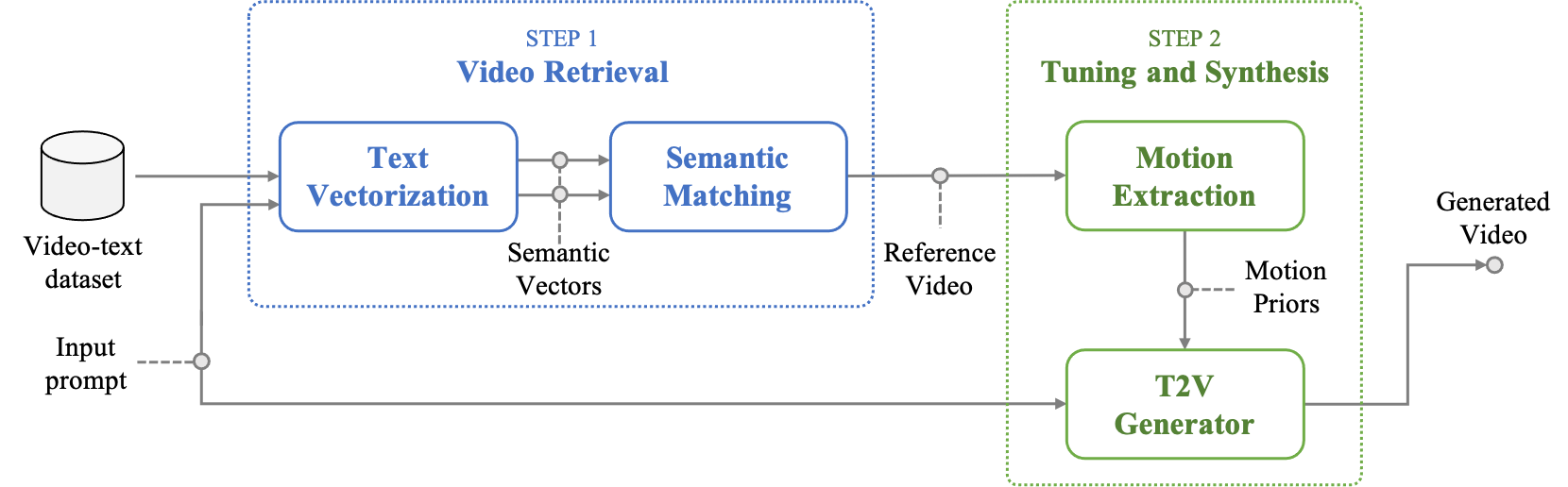} % Adjust the width as needed
  \caption{Pipeline overview. 
  This pipeline searches for videos with similar motion semantics, and extracts relevant information to fine-tune a pre-trained T2V model for video generation.
  }
  \label{fig_method}
\end{figure}

Given an input text prompt $P$ and a text-video dataset $\mathbb{D}=\{(T_i,V_i)\}^{N}_{i=1}$, 
our goal is to generate a high-quality video $V_{gen}$ by leveraging the rich dataset information. 
To achieve this, we propose a search-based pipeline, as shown in Fig. ~\ref{fig_method}, which consists of two steps:
\begin{itemize}
    \item \textbf{Video Retrieval.} For a given prompt input $P$, we first conduct text vectorization (Sec. ~\ref{method_vec}) to abstract it into semantic vectors. 
    Then, we conduct a matching process (Sec. ~\ref{method_search}).
    In this process, we utilize the extracted vectors to select a text-video pair from the dataset, which has the most similar motion semantics to the input prompt. The video of the selected text-video pair will be served as reference video $V_{r}$.

    \item \textbf{Tuning and Synthesis} (Sec. ~\ref{method_extract}). Once the referenced video $V_{r}$ is obtained, we perform motion extraction to get the most representative visual frames $\{f\}_n$ from $V_{r}$. Then we distill $\{f\}_n$ to fine-tune a pre-trained T2V model $G_{\theta}$. Finally, the input prompt $P$ is fed into the tuned generator $G_{\theta^{*}}$ to generate the final result $V_{gen}$.
\end{itemize}

%%%%%%%%%%%%%%%%%%%%%%%%%%%%%%%%%%%%%%%%%

%%%%%%%%%%%
\subsection{Text Vectorization}
\label{method_vec}

For a given text prompt $P$, we abstract it into two types of semantic vectors to represent its meaning. 
% propose two types of vectors to represent its semantics:

\paragraph{Sentence-level Semantic Vector.}
To represent the semantics of the whole sentence, we compute dense vector representations $\textbf{v}^{total}_{P}$ for $P$ utilizing a pre-trained language model~\cite{sentencebert}. 

\paragraph{Atomized Semantic Vectors.}
In order to represent the semantics of $P$ in finer granularity, 
remove irrelevant semantics (such as various adjectives and temporal states), 
and extract the core semantic structure, 
we abstract the text into atomized semantics vectors, 
which allows for customized semantic matching in subsequent steps.
Specifically, for the input text $P$, we decompose it into $m$ atomized semantic units \( S_P = \{s\}_m \). 
Each unit \( s = (\mathbf{v}^{mot}, \mathbf{v}^{atr}, \mathbf{v}^{rec}) \in S_P \)  
represents a distinct subject of the original semantic, 
which independently encapsulates a complete motion,
where the vectors denote the \textit{motion} ($\mathbf{v}^{mot}$), the \textit{actor} ($\mathbf{v}^{atr}$), and the \textit{recipient} ($\mathbf{v}^{rec}$), respectively.
It is worth mentioning that, in this process, we do not retain the semantics of adjectives, adverbs, and other such components in the original sentences. 
This is to make our semantic vectors focus on the motion itself and weaken the influence of appearance on motion matching, 
thus enabling more accurate retrieval in the subsequent matching process.
To implement this operation mentioned above,
 we first parse the original text $P$ into a \textit{dependency parse tree}~\cite{dependency, factorNLP, choi2015depends} to establish the relationships between each word. 
Then, we use verbs as cues to extract the corresponding subject and object word, forming multiple phrase groups. 
For each group, we use a pre-trained language model~\cite{spacy} to extract their word vectors to obtain semantic unit \( s \). 
Among all the units, we select the unit that contains the subject or predicate semantics of $P$ as the core unit $s_{\text{core}} \in S_P$, indicating it contains the essential dynamic semantics of the entire video.

%%%%%%%%%%%
\subsection{Semantic Matching}
\label{method_search}

After obtaining the prompt semantic vectors $(\mathbf{v}_P^{total}, S_P)$, 
we match them against the semantic vectors $(\mathbf{v}_i^{total}, S_i)$ of all texts in the dataset $\mathbb{D}$,
which are pre-computed and stored before the T2V process.
Our goal is to find a text-video pair $(T_{r},V_{r})\in\mathbb{D}$ as a reference, which maintains the most similar semantic and dynamic information with the input. The search process consists of two steps:

\paragraph{Coarse Filtering.} 
We first employ coarse filtering to narrow down the selection range.
We calculate the cosine similarity between the semantic vector \( \mathbf{v}^{total}_{P} \) of the input \( P \) and each semantic vectors \( \mathbf{v}^{total}_{i} \) of the dataset label $T_i \in\mathbb{D}$. 
Based on the computational results, we apply a coarse filtering to eliminate a batch of cases that do not meet the threshold $t_{total}$ condition.
Similarly, we subsequently filter by comparing the core semantic units \( s_{\text{core}} \) using \( \mathbf{v}^{mot}, \mathbf{v}^{atr}, \mathbf{v}^{rec} \).

\paragraph{Ranking.} 
After the initial filtering, we conduct a similarity-based ranking to retrieve the optimal text-video pair \((T_r, V_r)\). We design a ranking score $Score(T_i, P)$  to evaluate the motion semantic similarity among text \(T_i\) and prompt $P$, which can be calculated as follows:
\begin{equation}
\label{equ_ranking}
    Score(T_i, P) = sim(\mathbf{v}_{i}^{total}, \mathbf{v}_{P}^{total}) + F(S_i, S_P)
\end{equation}      
The \(sim(\cdot, \cdot)\) refers to cosine similarity. The matching score among semantic units $F$ is defined as :
\begin{equation}
\label{equ_ranking2}
    F(S_i, S_P) = \alpha \cdot sim(\mathbf{v}_{i,core}^{mot}, \mathbf{v}_{P,core}^{mot}) + 
        \beta \cdot sim(\mathbf{v}_{i,core}^{atr}, \mathbf{v}_{P,core}^{atr}) + 
        \gamma \cdot sim'(S_i, S_P)
\end{equation}
where \(\alpha\), \(\beta\), and \(\gamma\) are the coefficients of the respective parameters. And the \(sim'(\cdot, \cdot)\) is defined as:
\begin{equation}
    sim'(S_i, S_P) = \frac{1}{|S_P|} \sum_{s_p \in S_P} \max_{s_i \in S_i} 
        (sim(s_p^{mot}, s_i^{mot}) + sim(s_p^{atr}, s_i^{atr}) + sim(s_p^{rec}, s_i^{rec} ))
    % sim'(S_i, S_P) = \frac{1}{|S_P|} \sum_{s_p \in S_P} \max_{s_i \in S_i} sim\_comb(s_p, s_i),
    % sim\_comb(s_p, s_i) = sim(s_p^{act}, s_i^{act}) + sim(s_p^{actr}, s_i^{actr}) + sim(s_p^{rec}, s_i^{rec})
\end{equation}

% By adjusting the parameters \(\alpha\), \(\beta\), \(\gamma\), and coarse filtering threshold, we can emphasize the similarity of core motions.
It is worth mentioning that in many instances of motion information, 
the action is strongly coupled with the object of the action (for example, "\textit{excavator digging a hole}" and "\textit{mole digging a hole}" correspond to two completely different sets of dynamic information). 
Therefore, when calculating the similarity in Formula ~\ref{equ_ranking2}, 
we take the actor similarity into consideration.
Finally, we can obtain the optimal text-video pair:
% \((T_r, V_r)\) by maximizing \( Score(T_i, P) \):
\begin{equation}
    (T_{r},V_{r}) = \underset{(T_i,V_i)\in\mathbb{D}}{\arg\max} \, Score(T_i, P)
\end{equation}

%%%%%%%%%%%
\subsection{Motion Extraction and Video Synthesis}
% \subsection{Tuning and Synthesis}
\label{method_extract}

In this module, we extract the key frames $\{f\}_n$ from the retrieved video $V_r$ that best represent the information required by the prompt,
and distill them into priors to fine-tune the pre-trained generator.

\paragraph{Keyframe Extraction.}
As the videos from the text-video datasets are often lengthy and contain many irrelevant objects, we design a module to ensure that the visual information of extracted keyframes is concentrated on the text.
Given the input text $P$ and the retrieved text-video pair $(T_{r}, V_{r})$, 
we match their semantic units $S_r$ with input's $S_P$,
to find the subset $\{s_r\}' \subseteq S_r$ that closely matches elements in $S_P$. 
For each $s_{r,i} \in \{s_r\}'$, we take its original text and add it to the captions.
Next, for each element in the captions, 
we conduct open-set object detection~\cite{groundingdino} to detect the frames in video $V_x$ where the target appears and the approximate appearance bounding box. 
% Finally, based on the set of frames and the bounding boxes, we extract the final keyframe set $\{f\}_n$.
We integrate the detected results to identify the most critical time segments and the key image regions in the video. 
Based on this analysis, we finally crop and obtain the keyframe set $\{f\}_n$.

\paragraph{Motion Distillation.}
\label{method_dis}
After obtaining the keyframe set $\{f\}_n$, we employ \textit{Temporal Attention Adaptation} \cite{vmc} to extract motion information from $\{f\}_n$ and fine-tune the pre-trained T2V model $G_{\theta}$. The optimization process is achieved by minimizing the alignment loss function, as follows:

\begin{equation}
    \theta^{*} = \arg\min _\theta \mathbb{E}_{t, n, \boldsymbol{\epsilon}_t^n, \boldsymbol{\epsilon}_t^{n+1}}\left[\ell_{\cos }\left(\delta \boldsymbol{\epsilon}_t^n, \delta \boldsymbol{\epsilon}_{\theta, t}^n\right)\right],
\end{equation}

% \begin{equation}
% \ell_{\cos }(\boldsymbol{x}, \boldsymbol{y})=1-\frac{\langle\boldsymbol{x}, \boldsymbol{y}\rangle}{\|\boldsymbol{x}\|\|\boldsymbol{y}\|}
% \end{equation}

where $\ell_{\cos }(\boldsymbol{x}, \boldsymbol{y})=1-\frac{\langle\boldsymbol{x}, \boldsymbol{y}\rangle}{\|\boldsymbol{x}\|\|\boldsymbol{y}\|} $, and $n$ refers to the number of keyframes.
% and $c$ representing the fixed frame stride which remains fixed at $c=1$ across all experiments.
By conducting the optimization, 
we enhance the generator's capability to synthesize video to effectively capture and reproduce motion patterns required by the prompt $P$.
% The refined model effectively captures and reproduces complex motion patterns, resulting in improved video generation quality.

%%%%%%%%%%%
% \subsection{Video Synthesis}
\paragraph{Video Synthesis.}
\label{method_syn}
Using the fine-tuned model $ G_{\theta^{*}}$, 
we perform text-to-video synthesis based on the input prompt $P$ to generate the final video $V_{gen}$. 
This process can be formalized as $V_{gen} = G_{\theta^{*}}(P)$.

%%%%%%%%%%%%%%%%%%%%%%%%%%%%%%%%%%%%
%%%%%%%%%%%%%%%%%%%%%%%%%%%%%%%%%%%%
%%%%%%%%%%%%%%%%%%%%%%%%%%%%%%%%%%%%
%%%%%%%%%%%%%%%%%%%%%%%%%%%%%%%%%%%%
\section{Experiments}

In this section, we perform evaluations to validate the proposed method. 
We first describe the implementation details and dataset (Sec. ~\ref{exp_implement}), 
then compare the performance with existing methods (Sec. ~\ref{exp_perform}), 
and finally provide an analysis of the method design (Sec. ~\ref{exp_abu}).
We also present the generation results of the proposed method at Fig. ~\ref{fig:sample} and \textit{Webpage}.
% Sec.~\cite{exp_implement} introduces the employed datasets, Sec. 4.2 defines the evaluation
% protocol, Sec. 4.3 shows ablations of our diffusion framework and architectural choices, Sec. 4.4 quantitatively compares our method to state-of-the-art large-scale video generators and Sec. 4.5 performs qualitative evaluation. 
% We complement evaluation by showcasing samples in the Website.

\subsection{Implementation Details}
\label{exp_implement}

\textbf{In the video retrieval step}, we search for videos from the WebVid-10M~\cite{webvid} dataset.
which contains 10 million video clips with captions, sourced from the web. 
These videos are diverse and rich in their content. 
We utilize pre-trained language models from spaCy~\cite{spacy} and Sentence Transformers~\cite{sentencebert} to compute the word vectors of the prompt and video label.
The coarse filter parameters introduced in Sec. ~\ref{method_search} is set as $t_{total}=0.3, t_{mot}=0.9, t_{atr}={0.4}$. For each round of filtering, we have a fail-safe mechanism in place to ensure that at least the top ten highest-scoring cases are retained.
The ranking parameters is set as $\alpha = 1$, $\beta = 1$, $\gamma = 0.5$ in Eq. ~\ref{equ_ranking}.

\textbf{In the tuning and synthesis step}, we utilize Show-1~\cite{show1} as the pre-trained base T2V model, and use VMC~\cite{vmc} as the distill module. 
The proposed method generates a single video comprising 29 frames at a resolution of $576\times320$.
All experiments of our methods are conducted on a single NVIDIA RTX 4090 GPU with 24GB vRAM.

\subsection{Performance Evaluation}
\label{exp_perform}
% \subsubsection{Qualitative Comparison}
\begin{figure}[h]
  \centering
  \includegraphics[width=1\textwidth]{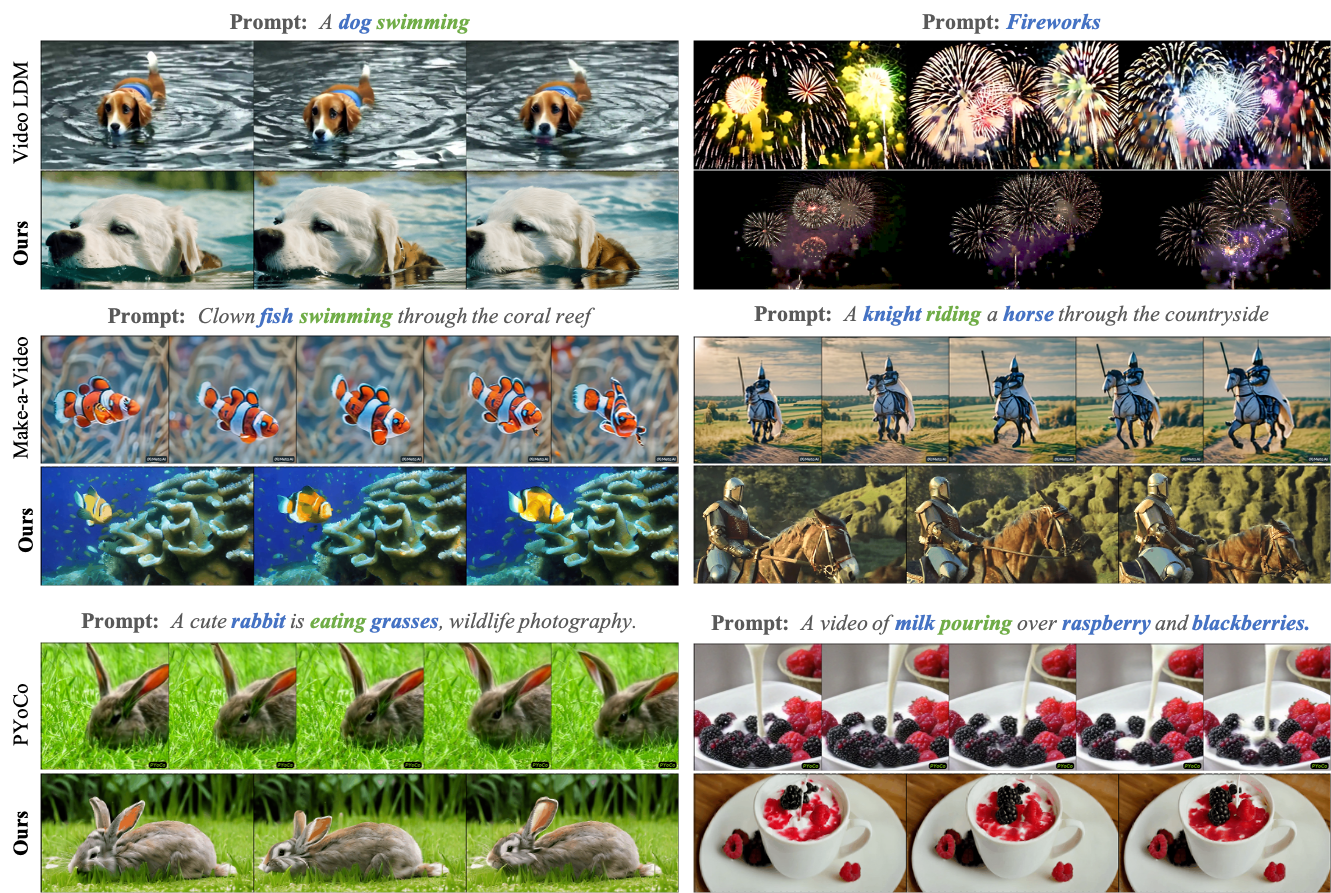} % Adjust the width as needed
  \caption{
      Qualitative comparison with state-of-the-art models. 
      Compared to Video LDM~\cite{videoldm} and Make-a-Video~\cite{makeavideo}, 
      our method generates more temporally coherent motion. 
      Compared to PyoCo~\cite{pyoco}, ours generates more detailed and realistic frames.
      Best view in \href{https://hrcheng98.github.io/Search_T2V/}{\textbf{\textcolor{blue}{\textit{Website}}}}.}
\label{fig_compare2}
\end{figure}

\begin{figure}[h]
  \centering
  \includegraphics[width=1\textwidth]{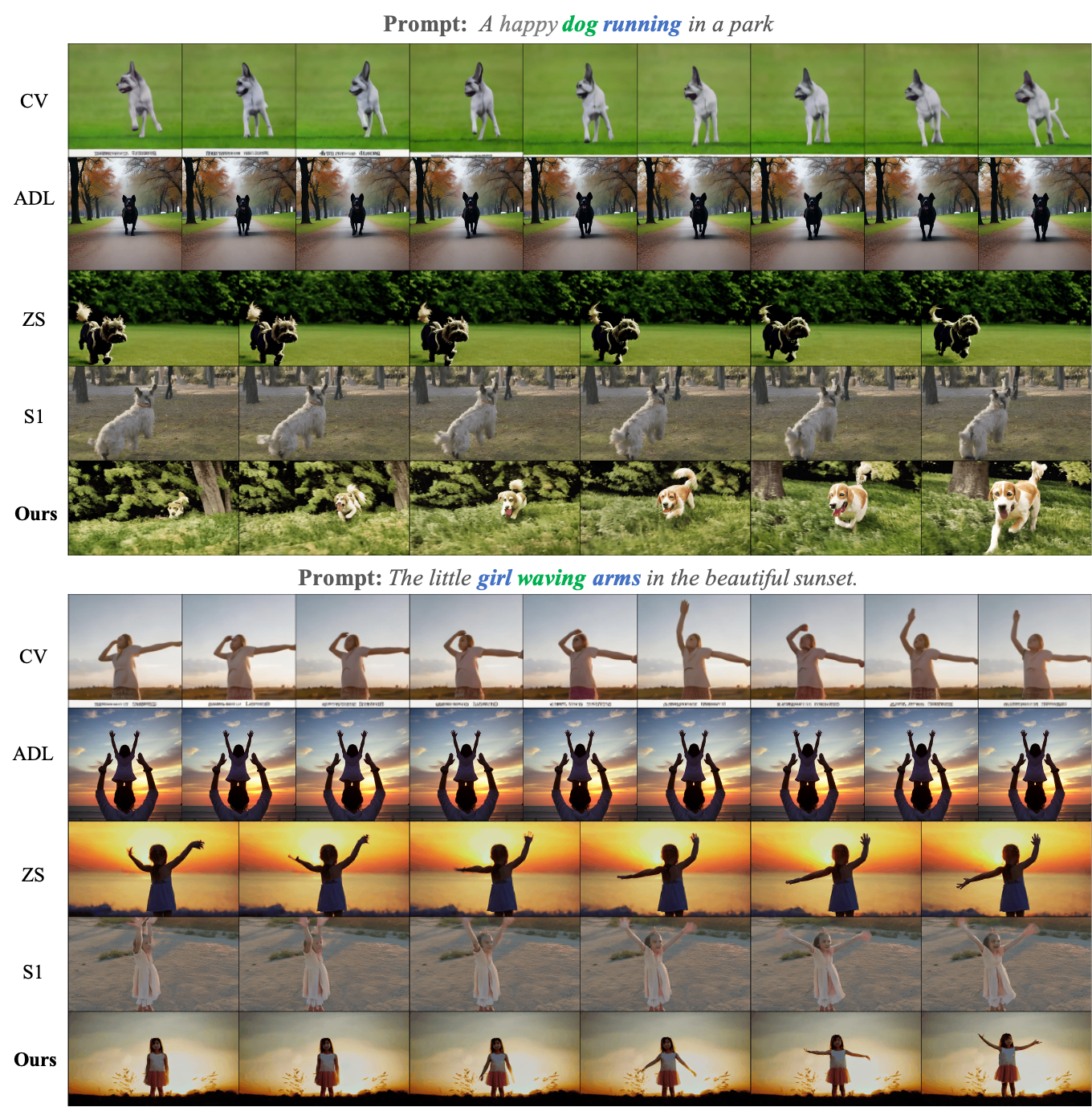} % Adjust the width as needed
  \caption{
    Qualitative comparison with existing T2V models. 
    All the results are generated by the officially released models.
    Compared to CV (abbr. for CogVideo~\cite{cogvideo}), ZS (abbr. for ZeroScope\_v2), and S1 (abbr. for Show-1~\cite{show1}), 
    our method generates more realistic appearances and more vivid motions. 
    For ADL (abbr. for AnimateDiff-Lightning~\cite{lin2024animatedifflightning}), 
    although it produces more detailed and realistic images, 
    it fails to accurately represent the motion information required by the text 
    (e.g., the dog is barely running, and the little girl's hand is almost static). 
    In contrast, our method generates videos with greater motion range and more dynamic realism.
    Best view in \href{https://hrcheng98.github.io/Search_T2V/}{\textbf{\textcolor{blue}{\textit{Website}}}}.
    }
\label{fig_compare1}
\end{figure}

\paragraph{Qualitative Comparison.}
% \paragraph{Subjective Results}
We present qualitative comparisons of our method against SOTA baselines on samples publicly released by the authors of Make-a-Video~\cite{makeavideo}, PYoCo~\cite{pyoco}, and Video LDM~\cite{videoldm}. 
Results are shown in Fig. ~\ref{fig_compare2}.
We also conduct comparison results against CogVideo~\cite{cogvideo},
Show-1~\cite{show1}, ZeroScope\_v2 \footnote{https://huggingface.co/cerspense/zeroscope\_v2\_576w},
and AnimateDiff-Lightning~\cite{lin2024animatedifflightning}. 
All the results are generated by the officially released models.
Results are shown in Fig. ~\ref{fig_compare1}.

% \subsubsection{User Study}
\paragraph{User Study.}
% photorealism, video-text alignment, motion quantity and quality
For human evaluation, we conduct a user study with 30 participants to assess three key aspects of generated samples, guided by the following questions:
(1) Visual Quality: \textit{How realistic is each static frame in the video?}
(2) Motion Quality: \textit{Is the video almost static? Are the dynamics consistent with common human understanding? Is the motion continuous and smooth?}
(3) Video-text Alignment: \textit{Does the video accurately reflect the target text? }
Each question is rated on a scale from 1 to 5, with higher scores indicating better performance.  
As shown in Tab. ~\ref{userstudy}, our method achieves the best human preferences on all evaluation parts.
For more details, please refer to Appendix ~\ref{apd_userstudy}.

\begin{table}
  \caption{
    \textbf{User study results} of different models: 
    Visual Quality (VQ), Motion Quality (MQ), and Video-text Alignment (VTA) ratings are on a scale from 1 to 5, with higher scores indicating better performance. 
    Our method achieves the highest scores across all evaluation criteria.
  }
  \label{userstudy}
  \centering
  \begin{tabular}{l|ccc|c}
    \toprule
     & \textbf{VQ} $\uparrow$ & \textbf{MQ} $\uparrow$ & \textbf{VTA} $\uparrow$ & \textbf{Avg. Score $\uparrow$} \\
    \midrule
    CogVideo~\cite{cogvideo} & 2.46 & 2.29 & 3.20 & 2.65 \\
    Show-1~\cite{show1}      & 3.15 & 2.94 & 3.88 & 3.32 \\
    ZeroScope\_v2           & 2.92 & 2.94 & 3.17 & 3.01 \\
    AnimateDiff-Lite~\cite{lin2024animatedifflightning}  & 3.23 & 2.93 & 3.43 & 3.20 \\
    \midrule
    Ours & \textbf{3.79} & \textbf{3.85} & \textbf{4.46} & \textbf{4.03} \\
    \bottomrule
  \end{tabular}
\end{table}

\subsection{Ablation Study}
\label{exp_abu}
\begin{figure}[h]
  \centering
  \includegraphics[width=1\textwidth]{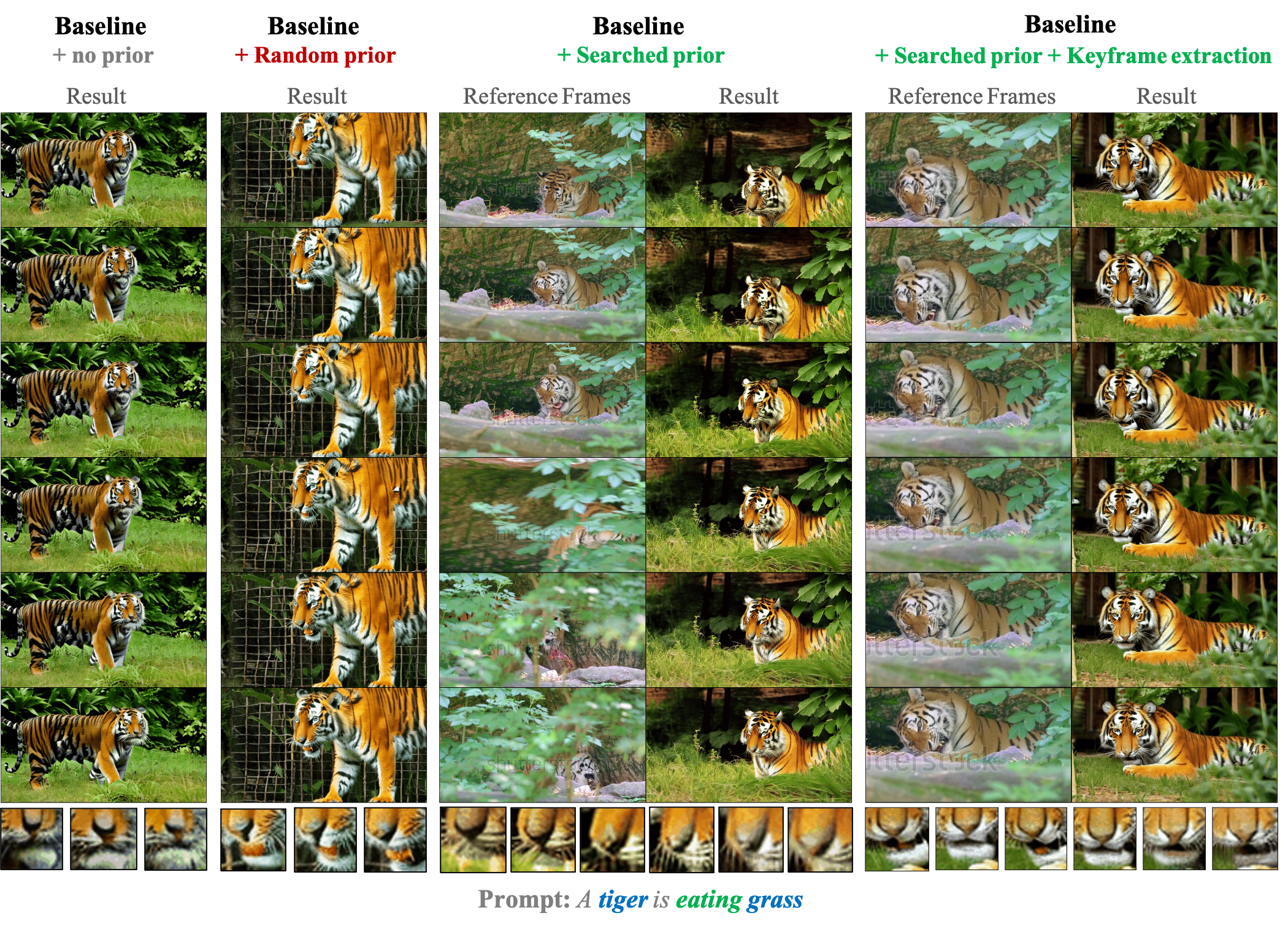} % Adjust the width as needed
  \caption{
    Comparison of different search and key frame extraction strategies. 
    We display a close-up of the tiger's mouth at the bottom of the image to better compare the \textit{"eating"} action.
    The full method obtains the best results.
    Best view in \href{https://hrcheng98.github.io/Search_T2V/}{\textbf{\textcolor{blue}{\textit{Website}}}}.}
  \label{fig:abu}
\end{figure}

\paragraph{Impact of video retrieval and motion extraction algorithm.} 
We perform several ablation experiments on different models of the proposed pipeline.
The generate result are presented in Fig. ~\ref{fig:abu}.
In the image, we can observe the following:
(1) Without adding any prior to the initial T2V pipeline, the video is almost static, and the tiger does not exhibit the eating action.
(2) When we add a prior to the model but remove the video retrieval module, 
the algorithm randomly chooses a video from the database as a reference.
%(e.g., the static train carriage scene shown). 
After distilling the dynamic information from this video and adding it as a prior, the model still fails to generate the eating motion.
(3) When we add a prior and enable the matching module, we obtain videos related to the desired action. 
However, if we do not use the keyframe extraction algorithm, 
the selected keyframes would contain a lot of irrelevant visual content 
(e.g. the tiger is obscured by leaves in the last three frames). 
This still does not help the model generate the desired motion.
(4) Finally, when we enable all modules, we obtain appropriate video keyframes. After adding this prior to the model, the generated tiger finally begins to chew the grass.

\paragraph{Impact of different sizes of the searching dataset.}
We validate the search results with datasets of different sizes. 
We set up six search datasets of varying sizes: 100\%, 50\%, 25\%, 10\%, 5\%, and 1\% of the original dataset~\cite{webvid} size, 
and conduct generation experiments using 10 identical prompts. 
The average scores calculated by Eq. ~\ref{equ_ranking} are shown in Tab. ~\ref{tab_datasetsize}. 
It can be observed that as the dataset size increases, the results obtained by the search become more ideal. However, the dataset size exhibits marginal effects. When the size of the original dataset reaches 50\% (i.e., 5 million text-video pairs), the search results approach the peak. Please refer to Appendix ~\ref{apd_ds_ablation} and Fig. ~\ref{apd_ds_ablation} for more experimental details and results (such as visualization of search results and generation results).

\begin{table}[h]
  \caption{Average ranking scores of different size datasets in search experiments.}
  \label{tab_datasetsize}
  \centering
  \begin{tabular}{c|cccccc}
    \toprule
    \textbf{Data size ($\times$ original size)} & \textbf{100\%} & \textbf{50\%} & \textbf{25\%} & \textbf{10\%} & \textbf{5\%} & \textbf{1\%} \\
    \midrule
    \textbf{Average Score} & 2.56 & 2.48 & 2.26 & 2.01 & 1.83 & 1.42 \\
    \bottomrule
  \end{tabular}
\end{table}

%%%%%%%%%%%%%%
% \section{Limitation Discussion}

\section{Conclusion}
\label{conclusion}

In this paper, we propose a novel approach for text-to-video generation that leverages existing real video resources as references during synthesis. 
% thereby enhancing the realism of generated videos. 
By integrating dynamic information from real videos as motion priors into a pre-trained T2V model, 
our method achieves high-quality video synthesis with precise motion characteristics at low costs. 
%This approach not only addresses the limitations of current video generation models but also provides a cost-effective solution by utilizing abundant real-world video resources available on the internet. 
This approach alleviates the issues of current video generation models and offers a cost-effective solution by leveraging the abundance of real-world video resources available on the internet. 
Our method offers the potential to popularize text-to-video generation by bridging the realism gap between synthetic and real-world motion dynamics, paving the way for applications in various domains such as entertainment, education, and virtual environments.

\paragraph{Limitation and Future Work} 
% Our proposed framework has great potential. 
% Some components in the current framework are just naive implementations. 
% There are many more ways to explore.
The proposed method has some limitations, including challenges in text-based video retrieval due to semantic ambiguities, difficulties in keyframe extraction missing broader dynamics, and the inability to learn abstract actions with the VMC~\cite{vmc} motion distillation method. 
Please refer to Appendix ~\ref{apd_fail} for more discussion on limitations.
Despite these issues, the approach shows promise for future exploration and improvement in this field.
In future work, it would be interesting to address the above limitations of our method.

% \medskip

\newpage

{
\small
\bibliography{main}
\bibliographystyle{plain}
}

\newpage

%%%%%%%%%%%%%%%%%%%%%%%%%%%%%%%%%%%%%%%%%%%%%%%%%%%%%%%%%%%%

\appendix

\section{Webpage}
The \textit{Webpage} of this project has been submitted in the supplementary materials. 
Since this is a video synthesis work, we \textbf{strongly recommend} to browse the videos to obtain a more intuitive judgment.

\section{Details of Ablation Analysis of Search Datasize}
\label{apd_ds_ablation}

\begin{figure}[h]
  \centering
  \includegraphics[width=1\textwidth]{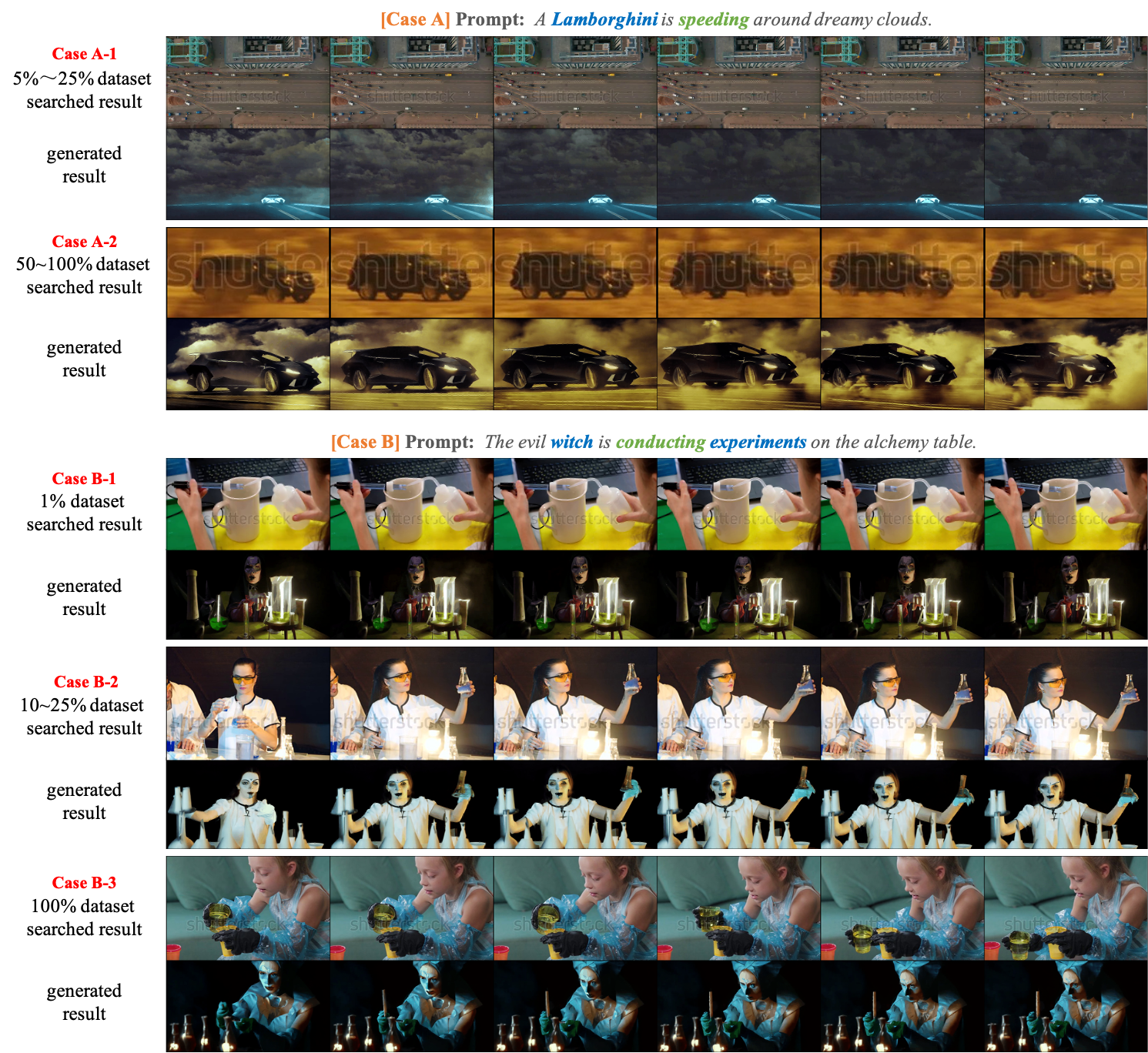} % Adjust the width as needed
  \caption{
    Samples of results generated by the T2V model tuned by different sizes of dataset. 
    % Please refer to \textit{Webpage file} for the best view and more samples.
    Best view in \href{https://hrcheng98.github.io/Search_T2V/}{\textbf{\textcolor{blue}{\textit{Website}}}}.
    }
  \label{fig:abu2}
\end{figure}

To verify the impact of different scales of search datasets on our method, 
we design this ablation experiment.
We set up search datasets with six different sizes: 100\%, 50\%, 25\%, 10\%, 5\%, and 1\% of the original dataset~\cite{webvid} size. 
Using 10 identical prompts, we conducted generation experiments. 
The prompt list is as follows. 
\begin{itemize}  
    \item \textit{An elephant is walking under the sea.  }
    \item \textit{The evil witch is conducting experiments on the alchemy table.  }
    \item \textit{A girl is holding an umbrella on a foggy street. }
    \item \textit{The train is heading towards the distance, at a station full of sakura blossoms. } 
    \item \textit{A Lamborghini is speeding down the highway in dreamy clouds.  }
    \item \textit{A panda rides a bicycle past the Eiffel Tower}  
    \item \textit{Flowers blooming in the spring garden.}  
    \item \textit{Fountains spraying water in a park.}  
    \item \textit{In the dark of night, the magnificent palace burns in flames.}  
    \item \textit{Scissors cutting through paper.}  
\end{itemize}

The search results and generation results are visualized in Fig.~\ref{fig:abu2}. 
From the figures, 
it can be seen that when the data scale is small, the proposed methods often fail to find suitable reference videos (e.g., in Case A-1, when the dataset is small, only overhead views of a racing track could be found; in Case B-2, the frame contains interference from another experimenter). 
Additionally, the retrieved videos from a small dataset always negatively impacts the subsequent performance of the generator (e.g., in Case B-1, the final generated video tends to be static). 
When the search dataset size increases, the retrieved videos significantly enhance the generated results (e.g. in Case A-2, the dynamics of the Lamborghini driving through the clouds are well presented).
% It can be conclude that as the dataset size increases, the videos retrieved become more closely related to the prompts. When the dataset size is around 50\%, it is sufficient to extract prior values.
% Therefore, we can conclude that the current dataset size can support the generation of videos for relatively common prompts.

%%%%%%%%%%%

\section{Limitation Discussion}
\label{apd_fail}

In our pioneering effort to reformulate the text-to-video framework into two distinct steps of video retrieval and tuning synthesis, we presents an implementation scheme. 
However, as this is the first exploration of this implementation format, limitations are inevitable. 
In this section, we discuss the limitation of the methods we found during the researching and developing process.

% \section{Limitation Discussion}
% \label{apd_fail}

\label{exp_abu}
\begin{figure}[h]
  \centering
  \includegraphics[width=1\textwidth]{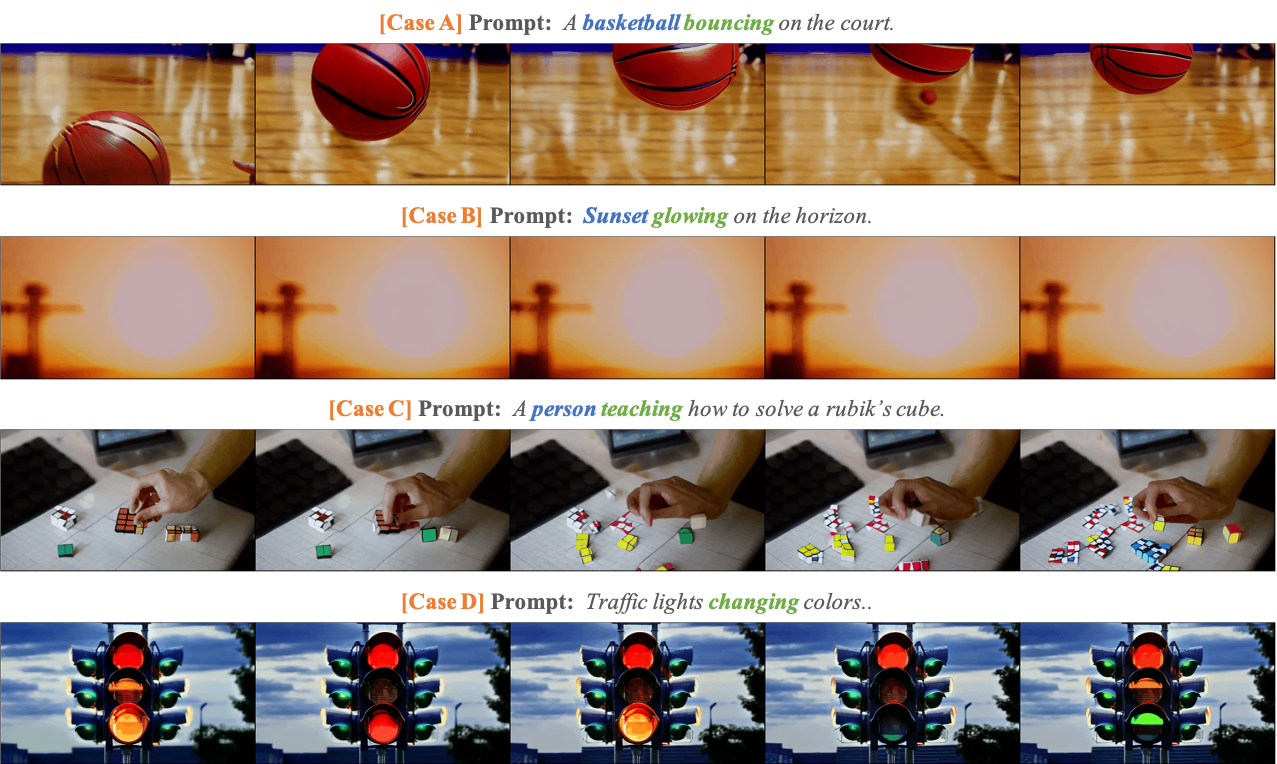} % Adjust the width as needed
  \caption{
    Samples of failed cases generated by proposed methods.
    In \textbf{Case-A}, due to the limitations of keyframe extraction, the proposed method only captures the bounding box of the basketball, but does not capture the interaction between the basketball and the hand. This results in the basketball being suspended in the air for a long time in the generated video. 
    In \textbf{Case-B}, similarly, after extracting the keyword sunrise, the proposed method only captures the features of the sun and its surrounding pixels, ignoring all the changes in the scenery due to the sunset, making the entire image inconsistent with the content required by the prompt. 
    In \textbf{Case-C}, due to the complexity and abstraction of teaching others to play Rubik's Cube, the model did not learn this motion information. 
    In \textbf{Case-D}, due to the complexity of the rules for traffic lights changing from red to yellow and then to green, 
    the generator faces difficulty in learning the rules, resulting in traffic lights flashing with incorrect rules.
    % Best view in \textit{Website} and \textit{Supplementary Video}.
    }
  \label{fig:fail}
\end{figure}

\paragraph{Limitations in the Retrieval Module}
Firstly, our text-based retrieval method inherently suffers from the limitations of textual meaning. At times, two vastly different texts can describe the same dynamic scene. For instance, "\textit{a car speeding over a viaduct}" and "\textit{bustling traffic in a thriving city}" might both refer to the same visual portrayal. This poses difficulties in text-based matching.
Moreover, the decoupling of motion and appearance from a semantic perspective is not always feasible. Take the action "\textit{dig}" as an example; the motion of a person digging differs significantly from that of a groundhog. Consequently, the coupling effect of motion and appearance during the search process can compromise the quality of priors.

\paragraph{Limitations in Keyframe Extraction} 
While keyframe extraction based on detection models is effective in most scenarios, it sometimes encounters difficulties.
For instance, a dynamic effect is often expressed through the entire frame's content, but detection models tend to focus on specific objects. In capturing keyframes for "\textit{sunset}" (as shown in Fig. ~\ref{fig:fail} Case-B), the algorithm might only capture the bounding boxes of the sun. The dynamic transitions of all objects from darkness to light are overlooked in this process. However, these transitions are integral to portraying the "\textit{sunset}" dynamic, yet the detection model disregards unrelated objects.

\paragraph{Learning of Abstract Dynamics}
Our framework also inherits limitations from the original VMC~\cite{vmc} motion distillation method. Due to its lightweight nature, it sometimes fails to learn abstract and complex actions like "\textit{sell}," "\textit{study}," or "\textit{enjoy}." These actions do not typically accompany explicit bodily movements, rendering them challenging for the current model.

Overall, though, this is a promising solution with numerous potential areas for further exploration. We eagerly anticipate future research endeavors in this direction.

%%%%%%%%

\section{Details of User Study}
\label{apd_userstudy}

\subsection{Objective}
To evaluate the performance of the proposed T2V method, we conduct this user study and compare the generated results with four classic open-source T2V method CogVideo~\cite{cogvideo}, Show-1~\cite{show1}, ZeroScope\_v2 and AnimateDiff-Lightning~\cite{lin2024animatedifflightning}.
The evaluation focuses on three key criteria: Visual Quality, Motion Quality, and Video-Text Alignment.

\subsection{Methodology}

\paragraph{Video Generation.} 
We select a set of 12 prompts to generate videos, which included dynamic content featuring people, animals, objects, and landscapes. 
For each prompt, videos are produced using five models. This process resulted in a total of 60 videos (12 prompts multiplied by 5 models).
% \begin{itemize}
%     \item A set of 12 prompts was created.
%     \item For each prompt, videos were generated using five different models: our video diffusion model and four other open-source models.
%     \item This resulted in a total of 60 videos (12 prompts $\times$ 5 models).
% \end{itemize}

% \paragraph{Participant Recruitment}
% \begin{itemize}
%     \item A diverse group of participants was recruited, ensuring a range of ages, backgrounds, and levels of expertise with video content.
%     \item Participants were recruited through online platforms and university mailing lists to ensure a broad demographic.
% \end{itemize}

\paragraph{Study Procedure.} 
    Each participant are asked to evaluate a series of videos. 
    For each video, participants should provide three separate scores 
    (each ranging from 1 to 5, with 1 being the lowest and 5 being the highest) 
    based on the following criteria:
    \begin{itemize}
        \item \textbf{Visual Quality:} \textit{How realistic is each static frame in the video? }
        \begin{itemize}
            \item 1 point: The content of the video is hard to discern.
            \item 3 points: The content of the video is clear but obviously synthesized.
            \item 5 points: The video is indistinguishable from real footage.
        \end{itemize}
        \item \textbf{Motion Quality:} 
        \textit{Is the video almost static? 
        Are the dynamics consistent with common human understanding? 
        Is the motion continuous and smooth?}
        \begin{itemize}
            \item 1 point: The video is nearly static, or movements are completely unnatural.
            \item 3 points: The video shows movement, but some aspects contradict common sense.
            \item 5 points: Movements are smooth and align with human expectations.
        \end{itemize}
        \item \textbf{Text-Video Alignment:}
        \textit{Does the video accurately reflect the target text?}
        \begin{itemize}
            \item 1 point: The video's content is completely unrelated to the prompt.
            \item 3 points: Either the objects or the actions in the video relate to the prompt.
            \item 5 points: The video's content fully matches the prompt's description.
        \end{itemize}
    \end{itemize}
    Participants are instructed to score each video independently and provide honest feedback.

\paragraph{Data Collection}
We collect test data using an online form.
Scores for each criterion are averaged across participants and videos. 
Statistical analysis is then conducted based on the data.

\subsection{Results Interpretation}
The results are shown in Tab. ~\ref{userstudy}.
Due to file size limitations for submissions, we are temporarily unable to publicly share the specific video examples and their scoring details involved in the user study. 

% \subsection{Ethical Considerations}
% \begin{itemize}
%     \item Participants were informed about the study's purpose and provided consent.
%     \item Data was anonymized to ensure privacy.
%     \item The study adhered to ethical guidelines for human-subject research.
% \end{itemize}

% \subsection{Conclusion}
% The user study aims to provide a comprehensive evaluation of the video diffusion model's performance in generating realistic and contextually accurate videos. The insights gained will guide further improvements and development of the model.

\section{Boarder Impact and Safeguards}
\label{apd_boarder}
% The proposed novel T2V format could enhance video realism in T2V synthesis and benefit research fields and industries such as entertainment, education, and content creation. On the other hand, this paper acknowledges potential negative societal impacts including the risk of generating highly realistic deepfakes, which could be used for disinformation, fake profile creation, or other malicious purposes. To mitigate these risks, the authors suggest implementing mechanisms for monitoring misuse and ensuring ethical guidelines are followed in the development and deployment of the technology.

In this section, we discuss the potential societal impacts of our proposed work and outline the safeguards to mitigate negative consequences.

The proposed novel T2V format could significantly enhance video realism in video synthesis, offering substantial benefits to research fields and industries such as entertainment, education, and content creation. This advancement promises to revolutionize these areas by enabling more immersive and realistic video content.
However, we also acknowledge the potential negative societal impacts of this technology. Notably, the ability to generate highly realistic deepfakes poses serious risks, including the potential for disinformation, the creation of fake profiles, and other malicious uses. To mitigate these risks, we suggest implementing robust mechanisms for monitoring misuse and ensuring that ethical guidelines are rigorously followed during the development and deployment of the technology.

% \section{More Results}
% \label{apd_more}

\end{document}